\documentclass[preprint,12pt]{elsarticle}

\usepackage{amssymb}

\usepackage[T1]{fontenc}
\usepackage{hyperref}

\usepackage{color}
\usepackage[cp1250]{inputenc}

\journal{Applied Soft Computing}

\begin{document}

\begin{frontmatter}


\title{A new algorithm for identity verification\\ based on the analysis of a handwritten\\ dynamic signature}


\author[IISI]{Krzysztof Cpa\l ka}
\ead{krzysztof.cpalka@iisi.pcz.pl}
\author[IISI]{Marcin Zalasi\'nski\corref{cor1}\fnref{fnmz}}
\ead{marcin.zalasinski@iisi.pcz.pl}
\author[IISI]{Leszek Rutkowski}
\ead{leszek.rutkowski@iisi.pcz.pl}
\address[IISI]{Institute of~Computational Intelligence, Cz\k{e}stochowa University of~Technology,\\Al. Armii Krajowej 36, 42-200 Cz\k{e}stochowa, Poland}
\cortext[cor1]{Corresponding author}
\fntext[fnmz]{Full postal address: Institute of~Computational Intelligence, Cz\k{e}stochowa University of~Technology, Al. Armii Krajowej 36, 42-200 Cz\k{e}stochowa, Poland; Telephone and fax number: 0-48 343250546}

\begin{abstract}
Identity verification based on authenticity assessment of~a~handwritten signature is an~important issue in biometrics. There are many effective methods for signature verification taking into account dynamics of~a~signing process. Methods based on partitioning take a very important place among them. In this paper we propose a~new approach to signature partitioning. Its most important feature is the~possibility of selecting and processing of~hybrid partitions in order to increase a~precision of~the~test signature analysis. Partitions are formed by a~combination of~vertical and horizontal sections of~the~signature. Vertical sections correspond to the initial, middle, and final time moments of~the~signing process. In turn, horizontal sections correspond to the signature areas associated with high and low pen velocity and high and low pen pressure on the~surface of a graphics tablet. Our previous research on vertical and horizontal sections of the dynamic signature (created independently) led us to develop the algorithm presented in this paper. Selection of~sections, among others, allows us to define the~stability of~the~signing process in the~partitions, promoting signature areas of~greater stability (and vice versa). In the~test of~the~proposed method two databases were used: public MCYT-100 and paid BioSecure.
\end{abstract}


\end{frontmatter}



\section{Introduction}
\label{sec:Introduction}

Security of~IT systems is related to, among others, effective identity verification of~system users. This verification may be performed using various methods based on: (a)~ something you have (e.g. chip card), (b)~something you know (e.g. password), (c)~something you are (e.g. biometric features). The~third approach is the~most convenient for people whose identity is verified and the~most difficult to forge for potential forgers. Therefore, it is very interesting and it creates possibilities for the~development of~new solutions. The~biometric features used in this approach are divided into two categories: (a)~physiological - related to the~construction of~the~human body (e.g. fingerprint, iris, hand geometry, face) and (b)~behavioural - related to the~human behaviour (e.g. signature, gait, keystrokes). A~handwritten signature occupies a special place among behavioural characteristics, its acquisition is not controversial and it is commonly socially acceptable.

In the~literature there are two main approaches to signature analysis. The~first one uses so-called static (off-line) signature and it is based on an~analysis of~geometric features of~the~signature, such as shape and size ratios, etc. (see e.g. \cite{Batista_2012,Radhika_2011,Kumar_2012,Beltran_2009,Carrera_2013,Guerbai_2015}). The~other approach is based on an~analysis of~the~dynamics of~signing process and it uses so-called dynamic (on-line) signature. Some authors have also presented a methods based on both these approaches (see \cite{Galbally_2015,Radhika_2015}). The~most commonly used signals, which are the~basis of~the~dynamic signature analysis, are pen pressure on the~tablet surface and pen velocity. Velocity signal is determined indirectly on the~basis of~pen position signals on the~tablet surface. The~dynamic signature verification is much more effective than the~static one because: (a)~dynamics of~signing is a very individual characteristic feature of~the~signer, (b)~it is difficult to forge, (c)~waveforms describing the~dynamics of~the~signature (even if you have them) are difficult to translate into the~process of~signing, but they are relatively easy to analyse. It should also be emphasized that the~algorithms for the~analysis of~the~dynamic signature can be relatively easily used in other areas of~application in the~field of~biometrics, which are based on the~analysis of~dynamic behaviour (see e.g.~\cite{Dean_2010, Ekinci_2007}).

In the~literature there are four main approaches to the~dynamic signature analysis: (a)~global feature-based approach (see e.g. \cite{Fierrez_2005, Lumini_2009, Nanni_2006, Nanni_Lumini_2006, Nanni_2005}), (b)~function-based approach (see e.g. \cite{Faundez_2007,Jain_2002,Jeong_2011,Kholmatov_2005,Maiorana_2010,Nanni_Lumini_2008}), (c)~regional approach (see e.g. \cite{Cpalka_Zalasinski_ESWA_2014, Cpalka_Zalasinski_PR_2014, Faundez_2011, Fierrez_2007, Huang_2003, Khan_2006, Pascual_2011, Zalasinski_2013}) and (d)~hybrid approach (see e.g. \cite{Moon_2010,Nanni_2010,Parodi_2014}). Among these approaches to the~dynamic signature analysis, the~methods based on signature regions are very interesting and effective. In the~literature in this field one can find, among others, new methods of~selection of~the~signature regions characteristic of the~signer, new ways of~interpretation of~these signature regions and new ways of~signature classification based on selected regions. Many authors use Hidden Markov Models (see e.g. \cite{Fierrez_2007}). Other authors propose ways of~classification adapted to their methods. In \cite{Huang_2003} signatures are segmented into strokes and for each of~them the reliability measure is computed on the~basis of~the~feature values which belong to the~current stroke. In \cite{Khan_2006} a~stroke-based algorithm that splits velocity signal into three bands has been proposed. This approach assumes that low and high-velocity bands of~the~signal are unstable, whereas the~medium-velocity band is useable for discrimination purposes. A~more detailed review of~the~literature on the~dynamic signature verification has been presented in our previous papers (see e.g. \cite{Cpalka_Zalasinski_ESWA_2014, Cpalka_Zalasinski_PR_2014}).

\begin{table}[!t]
\renewcommand{\arraystretch}{1.3}
\caption{Main characteristics of~the~algorithms for the~on-line signature verification based on regional approach (\textbf{f1}- Does the~method allow to select partitions of~signature associated with the~time of~signing process in order to increase accuracy of~signature verification? \textbf{f2}- Does the~method allow to select partitions of~the~signature associated with the~dynamics of~signing process in order to increase accuracy of~signature verification? \textbf{f3}- Does the~method focus on fast performance? \textbf{f4}- Does the~method evaluate the~stability of~the~signature in selected parts of~the~signature? \textbf{f5}- Does the~method take into account a~hierarchy of~selected parts of~the~signature in the~classification process? \textbf{f6}- Is a~given way of~classification interpretable?)}
\label{tab:Main_features}
\centering
\begin{tabular}{p{6.5cm}p{.7cm}p{.7cm}p{.7cm}p{.7cm}p{.7cm}p{.7cm}}
\hline \hline
Characteristic of~the~method& f1 & f2 & f3 & f4 & f5 & f6 \\\hline
Khan et al. \cite{Khan_2006} & no & \bf{yes} & no & \bf{yes} & no & no \\
Ibrahim et al. \cite{Ibrahim_2010} & no & \bf{yes} & no & \bf{yes} & no & no \\
Fierrez et al. \cite{Fierrez_2007} & \bf{yes} & no & no & no & no & no \\
Huang and Hong \cite{Huang_2003} & no & \bf{yes} & no & \bf{yes} & \bf{yes} & no \\
Fa\'undez-Zanuy, Pascual-Gaspar \cite{Faundez_2011} & \bf{yes} & no & \bf{yes} & no & no & no \\
Pascual-Gaspar et al. \cite{Pascual_2011} & \bf{yes} & no & \bf{yes} & no & no & no \\
Zalasi\'nski, Cpa\l ka \cite{Zalasinski_2012} & no & \bf{yes} & no & \bf{yes} & \bf{yes} & no \\
Cpa\l ka et al. \cite{Cpalka_Zalasinski_PR_2014} & no & \bf{yes} & no & \bf{yes} & \bf{yes} & \bf{yes} \\
Cpa\l ka, Zalasi\'nski \cite{Cpalka_Zalasinski_ESWA_2014} & \bf{yes} & no & no & \bf{yes} & \bf{yes} & \bf{yes} \\\hline
\bf{Our method} & \bf{yes} & \bf{yes} & no & \bf{yes} & \bf{yes} & \bf{yes} \\\hline
\hline
\end{tabular}
\end{table}

Our experience with different methods for the~dynamic signature verification based on the~regional approach induced us to prepare the~method presented in this paper. In \cite{Cpalka_Zalasinski_PR_2014} we propose a~method which determines the importance of~each time moment of~signing process individually for each signer. The~method takes into account stability of~signing in the~considered time moments. The~stability is determined using reference signatures. In our other paper i.e. \cite{Cpalka_Zalasinski_ESWA_2014} we propose a~method, which allows one to select some typical areas in the~signature of~the~user, created as a~result of an analysis of~pen velocity and pen pressure signals. These areas are associated with high and low pressure and high and low velocity. The~proposed methods work with high accuracy and they also have several other important advantages (determination of~weights of~areas, taking into account all the~regions and their weights in the~signature verification, signature classification based on the~fuzzy classifier), so we have decided to develop a~new method that would be a~combination of those two methods. Initial attempts to introduce this method are outlined in \cite{Zalasinski_2014}. However, the algorithm presented in the~mentioned paper required an~iterative determination of~the~so-called border of~inclusion of~reference signatures. In this paper we managed to eliminate this procedure and thus greatly simplify evaluation of~the~similarity of~test signatures to their~reference signatures. Thanks to this fact, the~system to evaluate the~similarity of~test signatures to their~reference signatures is a~full one-class classifier. Moreover, interpretation of~the~partitions selected by the~method described in \cite{Zalasinski_2014} is different from the~interpretation in the~algorithm proposed in this paper and the~partition analysis in the~mentioned method is less precise. Different interpretation of~the~partitions resulted in the~need to change the~work of~the~algorithm in every step. Thus, the~algorithm for signature verification proposed in this paper is a~new one, which has not been presented in the~literature so far.

The~following~features (see~Table~\ref{tab:Main_features}) distinguish the proposed method from the others:

\begin{list}{-}{}
\item It uses fuzzy sets and fuzzy systems theory in evaluation of~the~similarity of~test signatures to their~reference signatures. Character of~such similarity is not precise and it is difficult to describe it using the~classical theory of~sets and two-valued logic. In the~proposed method we used "high similarity" and "low similarity" fuzzy sets to describe similarity values (see \cite{Zadeh_1965}). Next, we formulated fuzzy rules and we used approximate inference. Thanks to this we obtained a~complete fuzzy system used in the~phase of~the~test signature verification. In the~rules description the~system takes into account the~weights of~importance of~selected partitions. 

\item It allows to interpret the~knowledge accumulated in the~system used for signature verification. These possibilities result from the~fact that: (a)~The construction of~the~fuzzy rules takes into account interpretability criteria of~the~clear fuzzy rules described in the~literature (e.g. in \cite{Gacto_2011}). (b)~For all signers we used consistent structure of~the~fuzzy classifier, in which only the~values of~the~rules' parameters change, but the~reasoning scheme remains unchanged. (c)~Input and output signals of~the~fuzzy classifier and the~parameters of~its rules have a~specific interpretation, referring to the~similarity of~test signatures to their~reference signatures. Thanks to this, the~parameters (i.e. the~parameters of~membership functions and importance weights of~the~rules) can be determined analytically and the~system does not require a~learning process.

\item It selects partitions of~the~signature which have the~following interpretation: high and low velocity in the~initial, middle and final time moments of~signing, high and low pressure in the~initial, middle and final time moments of~signing.

\item It determines values of~weights of~importance for each partition. Weights values are proportional to the~stability of~reference signatures in the~partitions. Thanks to this, the~proposed method uses all partitions in the~evaluation of~signature similarity (with varying intensity).

\item It bases on four types of~signals: a~shape signal of~the~trajectory $x$, a~shape signal of~the~trajectory $y$, a~pressure signal of~the~pen $z$ and a~velocity signal of~the~pen $v$. They are available as a~standard for graphics tablets: the first three of~them are acquired directly from the~graphics tablet and the~velocity is the first derivative of~a~signature trajectory. Various types of~tablets may have different sampling frequency, so in this case acquired signals are subject to~the~standard normalization procedure. In addition, the~signatures should be pre-processed using other standard methods to match their length, rotation, scale and offset.

\item It allows to flexibly adjust a set of~signals describing the~dynamics of~the~signature to specific areas of~application and hardware capabilities. There are two most common variants of~the~method. The~first assumes that a graphics tablet is used in the~training phase and in the~verification phase. In this case the~precision of~the~proposed method is the~highest, because the~signals describing not only the~shape of~the~signature, but also its dynamics, are used in both phases. The~second variant assumes that in the~training phase the~graphics tablet is available and in the~verification phase we have a~stand-alone (not connected to the~computer) device with a~touch screen (e.g. a~smartphone, a~tablet, etc.), from which it is impossible to obtain information about the~pen pressure. In this case, the~partitions are determined in the~training phase on the~basis of~velocity and pressure. Verification phase takes into account only the~shape of~the~signature. In the~description of~the~method we took into account the~second variant, assuming that in the~signature verification phase the~signals describing the~dynamics of~the~signature may not be available. Obviously, in practice there may be also indirect modes of~action (e.g. based on the~generation of~velocity trajectories in the~training phase without knowing the~pressure trajectory or using the~angle of~the~pen to the~tablet surface during the~signing process), but the~proposed method can be adapted to each of~them.
\end{list}

\begin{figure}[!pth]
\centering
\includegraphics[scale=1.00, clip]{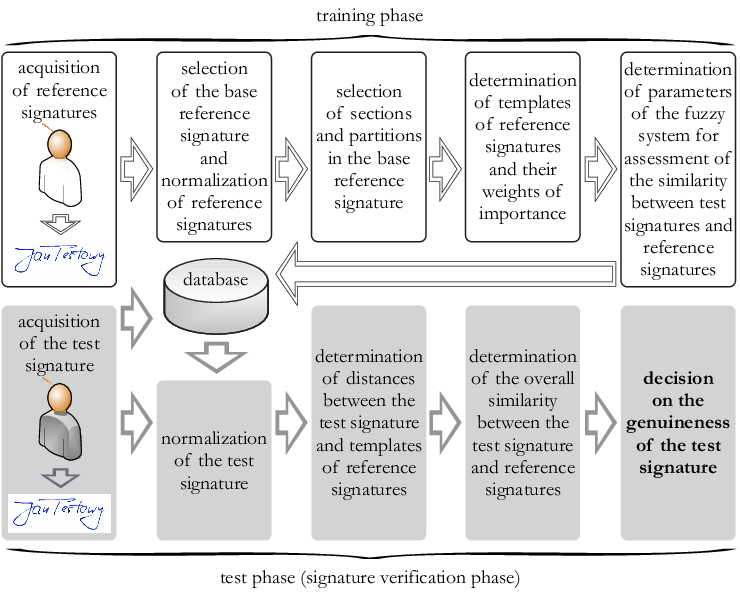}
\caption{A block diagram of~the~new algorithm for the~dynamic signature verification based on hybrid partitioning.}
\label{fig:Algorithm}
\end{figure}

In summary, the proposed new algorithm for identity verification based on the analysis of a handwritten dynamic signature is mainly characterized by new definition of the dynamic signature features (so-called hybrid partitions of the dynamic signature), new way of their processing and use of authorial one-class neuro-fuzzy classifier, whose structure is determined individually for each user without using forged signatures. Due to the above-mentioned mechanisms, the proposed algorithm works very precisely and individually for each user.

To test the~proposed method we used two databases of~the~signatures: public MCYT-100 (see \cite{MCYT}) distributed by the~Biometric Recognition Group - ATVS (see \cite{ATVS}) and paid BioSecure (BMDB) distributed by the~BioSecure Association (see \cite{www_BS}).

This paper is organized into 4 sections. Section~\ref{sec:Detailed_description_of_the_algorithm} contains a detailed description of~the~algorithm. The simulation results are presented in Section~\ref{sec:Simulation_results}. The conclusions are drawn in Section \ref{sec:Conclusions}.

\section{Detailed description of~the~algorithm}
\label{sec:Detailed_description_of_the_algorithm}

\begin{figure}[!pth]
\centering
\includegraphics[scale=0.83, clip]{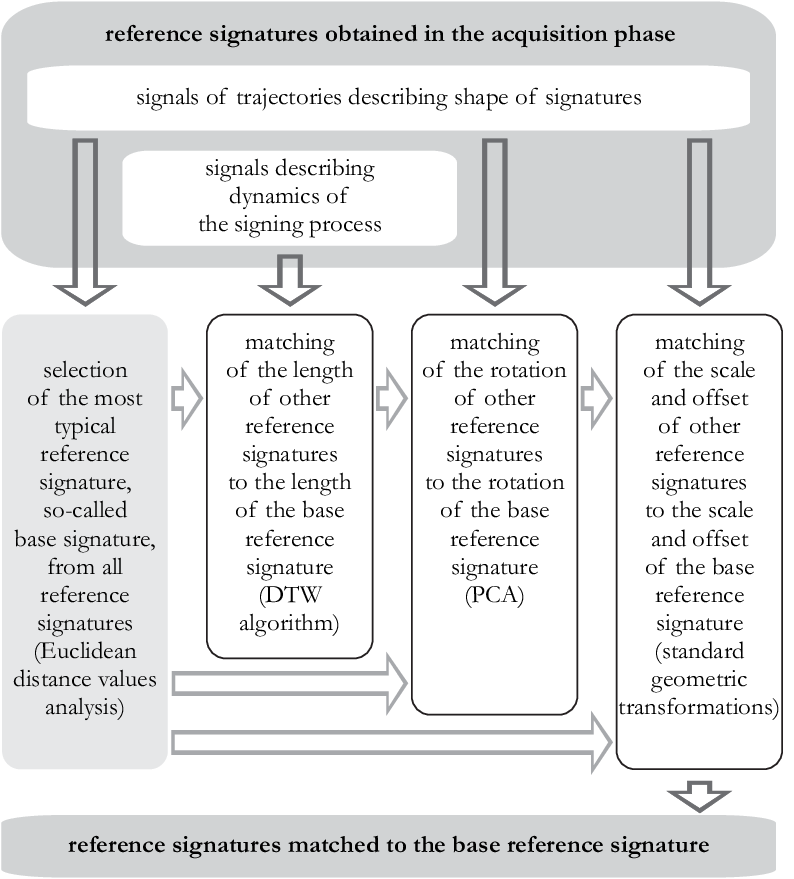}
\caption{Signature normalization procedure.}
\label{fig:PreProcessing}
\end{figure}

The proposed algorithm for the~dynamic signature verification based on hybrid partitioning works in two phases (see Fig.~\ref{fig:Algorithm}): the training phase (Section~\ref{sec:Learning_phase}) and the test phase (Section~\ref{sec:Testing_phase}). In both of~them a~procedure of~signature normalization is performed (see Fig.~\ref{fig:PreProcessing}). In this procedure for each user the~most typical reference signature, called base signature, is selected. It is one of~the~reference signatures collected in the~acquisition phase, for which the~distance to the~other reference signatures is the~smallest. The~distance is calculated according to the~adopted distance measure (e.g. Euclidean). Training or test signatures are matched to the~base signature using the~Dynamic Time Warping algorithm (see e.g. \cite{Banko_2012,Lee_2012,Su_2014}), which operates on the~basis of~matching velocity and pressure signals. The~result of~matching the~two signatures is a~map of~their corresponding points. On the~basis of~the~map, trajectories of~the~signatures are matched. Matching by way of using DTW could not be done directly with the~use of~trajectories, because this would remove the~differences between the~shapes of~the~signatures. It would have a~very negative impact on training. Elimination of~differences in rotation of~the~signatures is performed by the~PCA algorithm which in the~literature is commonly used to make image rotation invariant (see e.g. \cite{Gonzalez_2002}). The~scale and offset are compensated by standard geometric transformations. Various normalization techniques are described in detail in the~literature, so their description will not be included in this paper (see e.g. \cite{Jain_2002, Ibrahim_2010, Lei_2005, OReilly_2009}).

\subsection{Training phase}
\label{sec:Learning_phase}

At the~beginning of~the~training phase partitioning of~the~reference signatures of~each user is realized (Section~\ref{sec:Partitioning}). Partitions are hybrid because they are created from a~combination of~vertical and horizontal sections. Vertical sections are time intervals indicating the~initial, middle and final phase of~signing. Horizontal sections are created in each vertical section. In this process signals describing the dynamics of~a~signature are taken into account. If the~velocity signal is partitioned as first (order of~signal processing is arbitrary), then in each time interval the~average value of~the~velocity is determined (averaging discrete values of~the~velocity). Next, in each time interval two partitions are created: (a)~the one associated with the~velocity lower than the~average and (b)~the other one associated with the~velocity higher than the~average. The~procedure is analogous for the~second available signal - pen pressure. As a~result, the~following partitions are created: high and low velocity at the~initial, middle and final moment of~signing, high and low pen pressure on the~graphics tablet surface at the~initial, middle and final moment of~signing. The~number of~time moments which affects the~number of~partitions is a~parameter of~the~algorithm, which may be greater than or equal to 1. Each partition is physically a~subset of~points of~the~trajectory $x$ or $y$, which describes the~shape of~a~signature. After creation of~the~partitions, determination of~the~templates of~the~reference signatures is performed. Each template is associated with a~separate partition (Section~\ref{sec:Generation_of_the_templates}). With the~reference signature templates, the values of partition importance weights can be determined (Section~\ref{sec:Wyznaczenie_parametrów_elastycznego_klasyfikatora_rozmytego}). For example, the~weight of~importance of~the~first partition depends on the~similarity of~the~points of~the~reference signatures from the~first partition to the~corresponding components of~the~first template. This similarity is dependent on the~value of~the~Euclidean distance. Higher value of~the~partition importance weight means that a~specified part of~the~reference signatures associated with this partition was created in a~more stable way (with similar value of~the~velocity or the~pressure). Moreover, a~greater value of~the~weight means that in the~test phase fragments of~the~test signatures associated with this partition will be more important in evaluation of~the~similarity of~the~test signature to the~reference signatures. After creation of~the~partitions of~the~reference base signature, creation of~the~reference signatures templates and calculation of~the~weights of~importance, parameters of~the~flexible fuzzy one-class classifier are determined in the~learning phase (Section~\ref{sec:Wyznaczenie_parametrów_elastycznego_klasyfikatora_rozmytego}). The~classifier is used in the~test phase to evaluate the~similarity of~the~test signature to the~reference signatures. Obviously, partitions of~the~base signature, the~templates of~the~reference signatures, the~weights of~importance and the~parameters of~the~classifier are determined individually for each user and they must be stored in a~database.

\subsubsection{Creation of~the~partitions}
\label{sec:Partitioning}

Each reference signature $j$ ($j=1,2, \ldots,J$, where $J$ is the~number of~reference signatures) of~the~user $i$ ($i=1,2, \ldots,I$, where $I$ is the~number of~users) is represented by the~following signals:

\begin{list}{-}{}
\item \textbf{Signals describing the~shape of~a~signature.} Signal ${{\bf{x}}_{i,j}} = \left[ {{x_{i,j,k = 1}},} \right.$ ${x_{i,j,k = 2}}, \ldots,$ $\left. {{x_{i,j,k = {K_i}}}} \right]$~describes the~movement of~the~pen in the~two-dimensional space along the~$x$ axis, where ${K_i}$ is the~number of~signal samples. The number of~samples ${K_i}$ results from the~sampling frequency of~the~graphics tablet and performance of~the~DTW algorithm in the~normalization phase. Thanks to the~signature normalization, all trajectories describing the~signatures of~the~user $i$ have the~same number of~samples ${K_i}$. Movement of~the~pen along the~$y$ axis can be described in a similar way: ${{\bf{y}}_{i,j}}=\left[ {{y_{i,j,k=1}},{y_{i,j,k=2}}, \ldots,{y_{i,j,k={K_i}}}} \right]$. In order to simplify the~description of~the~algorithm we used the~same symbol ${{\bf{a}}_{i,j}}=\left[ {{a_{i,j,k=1}},{a_{i,j,k=2}}, \ldots,{a_{i,j,k={K_i}}}} \right]$ to describe both shape trajectories, where $a \in \left\{{x,y} \right\}$.

\item \textbf{Signals describing the~dynamics of~a~signature.} Signal ${{\bf{v}}_{i,j}}=\left[ {{v_{i,j,k=1}},{v_{i,j,k=2}}, \ldots,{v_{i,j,k={K_i}}}} \right]$ describes velocity of~the~pen and trajectory ${{\bf{z}}_{i,j}}=\left[ {{z_{i,j,k=1}},{z_{i,j,k=2}}, \ldots,{z_{i,j,k={K_i}}}} \right]$ describes the~pen pressure on the~surface of~the~graphics tablet. In order to simplify the~description of~the~algorithm we used the~same symbol ${{\bf{s}}_{i,j}}=\left[ {{s_{i,j,k=1}},{s_{i,j,k=2}}, \ldots,{s_{i,j,k={K_i}}}} \right]$ to describe both dynamics signals, where $s \in \left\{{v,z} \right\}$.
\end{list}

\begin{figure}[!pth]
\centering
\includegraphics[scale=1.00, clip]{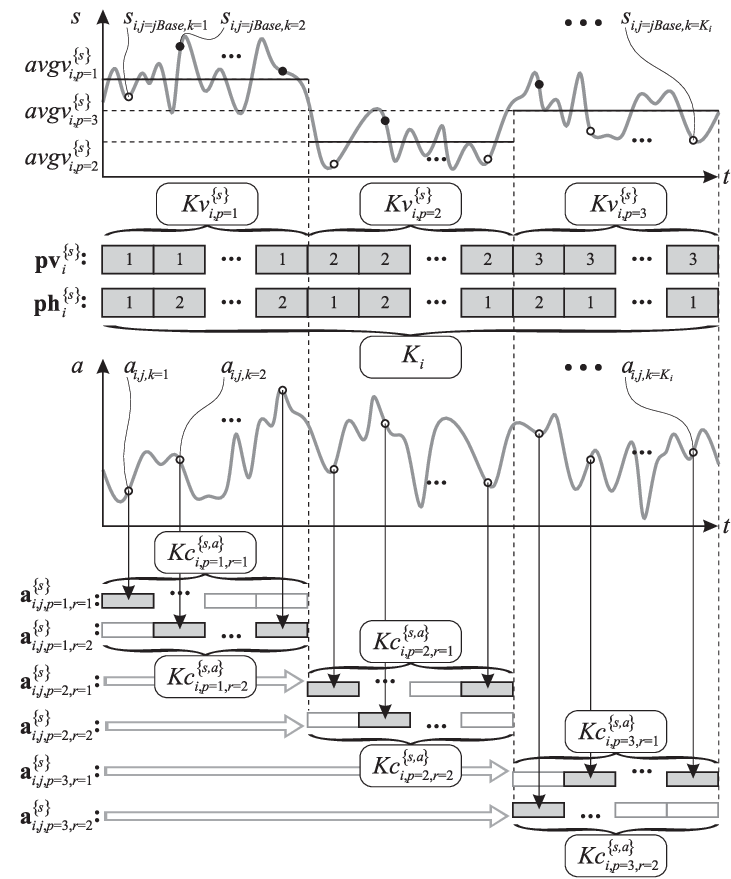}
\caption{Method for partitioning the~base reference signature $j=jBase$ on the~basis of~the~signal $s \in \left\{{v,z} \right\}$ and an~example of~partitioning the~trajectory $a \in \left\{{x,y} \right\}$ of~the~reference signature $j$ of~the~user $i$ for the~case ${P^{\left\{s \right\}}}=3$ and ${R^{\left\{s \right\}}}=2$. Method of~determination of~sections for different combination of~the~values ${P^{\left\{s \right\}}}$ and ${R^{\left\{s \right\}}}$ is realized analogously.}
\label{fig:Partitioning}
\end{figure}

The purpose of~the~partitioning is to assign each point of~the~signal ${{\bf{v}}_{i,jBase}}$ and the~signal ${{\bf{z}}_{i,jBase}}$ of~the~reference base signature to a~single hybrid partition, resulting from a~combination of~the~vertical and the~horizontal section, where $jBase \in \left\{{1, \ldots,J} \right\}$ is an~index of~the~base signature. As already mentioned, the~base signature is taken into account during partitioning as the~most typical reference signature of~the~user $i$. Therefore, it is selected from a~set of~reference signatures and it is not generated by averaging or grouping signals describing reference signatures.

The idea of~partitioning is shown in Fig.~\ref{fig:Partitioning}. At the~beginning of~the~partitioning, the~vertical sections of~the~signals ${{\bf{v}}_{i,jBase}}$ and ${{\bf{z}}_{i,jBase}}$ are created. Each of~them represents a different time moment of~signing: (a)~initial or final for the~case ${P^{\{s\}}}=2$, (b)~initial, middle or final for the~case ${P^{\{s\}}}=3$, (c)~initial, first middle, second middle or final for the~case ${P^{\{s\}}}=4$. The~vertical sections are indicated by the~elements of~the~vector ${\bf{pv}}_i^{\left\{s \right\}}=\left[ {pv_{i,k=1}^{\left\{s \right\}},pv_{i,k=2}^{\left\{s \right\}}, \ldots,pv_{i,k={K_i}}^{\left\{s \right\}}} \right]$ determined as follows:

\begin{equation}
\label{eq_pv}
pv_{i,k}^{\left\{s \right\}}=\left\{{\begin{array}{*{20}c}1 & {{\mathop{\rm for}\nolimits}} & {0 < k \le \frac{{K_i}}{{P^{\{s\}}}}} \\
2 & {{\mathop{\rm for}\nolimits}} & {\frac{K_i}{{P^{\{s\}}}} < k \le \frac{{2K_i}}{{P^{\{s\}}}}} \\
{} & \vdots & {} \\
{P^{\{s\}}} & {{\mathop{\rm for}\nolimits}} & {\frac{{(P^{\{s\}}-1 )K_i}}{{P^{\{s\}}}} < k \le K_i} \\
\end{array}}, \right.
\end{equation}

\noindent where $s \in \left\{{v,z} \right\}$ is the~signal type used for determination of~the~partition (velocity $v$ or pressure $z$), $i$ is the~user index ($i=1, 2, \ldots, I$), $j$ is the~reference signature index ($j=1, 2, \ldots, J$), $K_{i}$ is the~number of~samples of~normalized signals of~the~user $i$ (divisible by ${P^{\{s\}}}$), $k$ is an~index of~the~signal sample ($k=1, 2, \ldots, K_{i}$) and $P^{\{s\}}$ is the~number of~the~vertical sections ($P^{\{s\}}\ll{K_i}$ and ${P^{\left\{s \right\}}}={P^{\left\{v \right\}}}={P^{\left\{z \right\}}}$). A
number of~the~vertical sections can be arbitrary, but its increase does not increase interpretability and accuracy of~the~method (see Section~\ref{sec:Simulation_results}).

After creation of~the~vertical sections of~the~signals ${{\bf{v}}_{i,jBase}}$ and ${{\bf{z}}_{i,jBase}}$, horizontal sections are created. Each of~them represents high and low velocity and high and low pressure in individual moments of~signing. Horizontal sections indicated by the~elements of~the~vector ${\bf{ph}}_i^{\left\{s \right\}}=\left[ {ph_{i,k=1}^{\left\{s \right\}},ph_{i,k=2}^{\left\{s \right\}}, \ldots,ph_{i,k={K_i}}^{\left\{s \right\}}} \right]$ are determined as follows:

\begin{equation}
\label{eq_ph}
ph_{i,k}^{\left\{s \right\}}=\left\{{\begin{array}{*{20}{c}}
1&{{\rm{for}}}&{{s_{i,j=jBase,k}} < avgv_{i,p=pv_{i,k}^{\left\{s \right\}}}^{\left\{s \right\}}}\\
{\rm{2}}&{{\rm{for}}}&{{s_{i,j=jBase,k}} \ge avgv_{i,p=pv_{i,k}^{\left\{s \right\}}}^{\left\{s \right\}}}
\end{array}} \right.,
\end{equation}

\noindent where $jBase$ is the~base signature index, $avgv_{i,p}^{\left\{s \right\}}$ is an~average velocity (when $s=v$) or an~average pressure (when $s=z$) in the~section indicated by the~index $p$ of~the~base signature $jBase$:

\begin{equation}
\label{eq_avgv}
avgv_{i,p}^{\left\{s \right\}}=\frac{1}{{K{v_{i,p}}}}\sum\limits_{k=\left( {\frac{{\left( {p - 1} \right) \cdot {K_i}}}{{{P^{\{s\}}}}} + 1} \right)}^{k=\left( {\frac{{p \cdot {K_i}}}{{{P^{\{s\}}}}}} \right)} {{s_{i,j=jBase,k}}},
\end{equation}

\noindent where $K{v_{i,p}}$ is the~number of~samples in the~vertical section $p$, ${s_{i,j=jBase,k}}$ is the~sample $k$ of~the~signal $s \in \left\{{v,z} \right\}$ describing dynamics of~the~signature.

As a~result of~partitioning, each sample ${v_{i,jBase,k}}$ of~the~signal ${{\bf{v}}_{i,jBase}}$ of~the~base signature $jBase$ and each sample ${z_{i,jBase,k}}$ of~the~signal ${{\bf{z}}_{i,jBase}}$ of~the~base signature $jBase$ is assigned to the~vertical section (assignment information is stored in the~vector ${\bf{pv}}_i^{\left\{s \right\}}$) and horizontal section (assignment information is stored in the~vector ${\bf{ph}}_i^{\left\{s \right\}}$). The~intersection of~the~sections forms the~partition. Fragments of~the~shape trajectories ${{\bf{x}}_{i,j}}$ and ${{\bf{y}}_{i,j}}$, created by taking into account ${\bf{pv}}_i^{\left\{s \right\}}$ and ${\bf{ph}}_i^{\left\{s \right\}}$, will be denoted as ${\bf{a}}_{i,j,p,r}^{\{s\}}=\left[ {a_{i,j,p,r,k=1}^{\{s\}},a_{i,j,p,r,k=2}^{\{s\}}, \ldots,a_{i,j,p,r,k=Kc_{i,p,r}^{\left\{{s,a} \right\}}}^{\{s\}}} \right]$. The~number of~samples belonging to the~partition $\left( {p,r} \right)$ (created as an~intersection of~the~vertical section $p$ and the~horizontal section $r$, included in the~trajectory ${\bf{a}}_{i,j,p,r}^{\{s\}}$) of~the~user $i$ associated with the~signal $a$ ($x$ or $y$) and created on the~basis of~the~signal $s$ (velocity or pressure) will be denoted as $Kc_{i,p,r}^{\left\{{s,a} \right\}}$. It should be noticed that $\sum\limits_{p=1}^{{P^{\left\{s \right\}}}} {\sum\limits_{r=1}^{{R^{\left\{s \right\}}}} {Kc_{i,p,r}^{\left\{{s,a} \right\}}}}={K_i}$ for $a \in \left\{{x,y} \right\}$ and $s \in \left\{{v,z} \right\}$. The~number of~partitions of~the~base signature of~the~user $i$ is equal to ${P^{\left\{v \right\}}} \cdot {P^{\left\{z \right\}}} \cdot 4$. The~partitions are used to determine the~reference signature templates.

\subsubsection{Generation of~the~templates}
\label{sec:Generation_of_the_templates}

\begin{figure}[!pth]
\centering
\includegraphics[scale=1.00, clip]{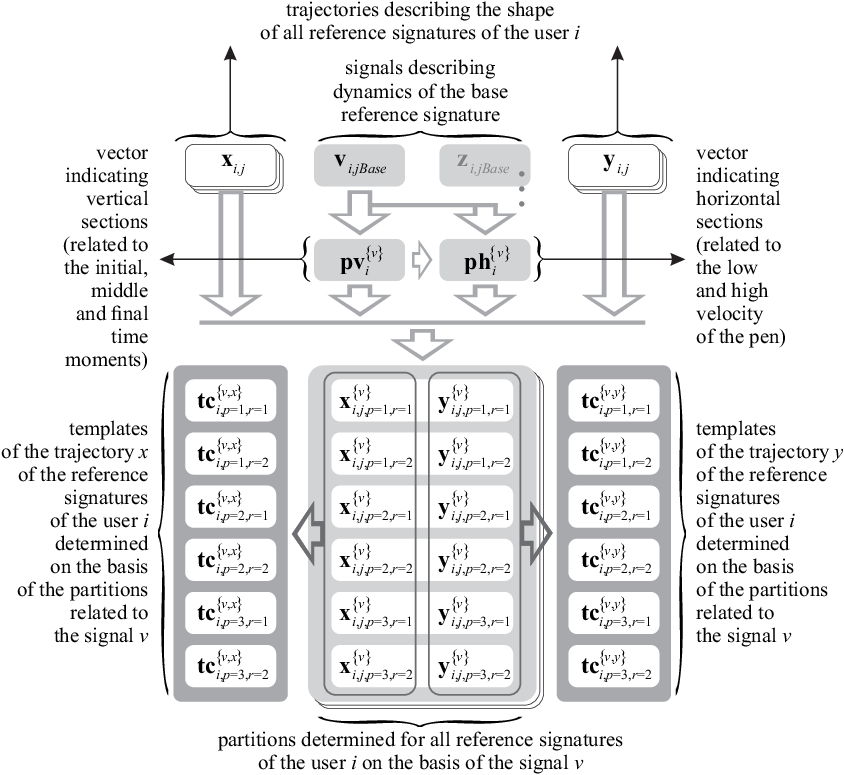}
\caption{Method of~determination of trajectories $x$ and $y$ of~the~templates of~the~user $i$ on the~basis of~the~partitions associated with the~signal $v$ for ${P^{\left\{s \right\}}}=3$ and ${R^{\left\{s \right\}}}=2$. Determination of~the~templates on the~basis of~the~partitions associated with the~signal $z$ and different values of~parameters ${P^{\left\{s \right\}}}$ and ${R^{\left\{s \right\}}}$ is realized in a similar way.}
\label{fig:Templates}
\end{figure}

The following parameters are considered when determining reference signature templates: (a)~all $J$ reference signatures of~the~user $i$, (b)~two shape trajectories of~the~reference signatures, i.e. ${{\bf{x}}_{i,j}}$ and ${{\bf{y}}_{i,j}}$ and (c)~partitions created for the~reference base signature, resulting from the~intersection of~the~vertical sections (indicated by ${\bf{pv}}_i^{\left\{s \right\}}$) and horizontal sections (indicated by ${\bf{ph}}_i^{\left\{s \right\}}$) (Fig.~\ref{fig:Templates}). The~templates of~the~signatures are averaged fragments of~the~reference signatures represented by the~shape trajectories ${{\bf{x}}_{i,j}}$ or ${{\bf{y}}_{i,j}}$. The~number of~the~templates created for the~user $i$ is equal to the~number of~the~partitions. Each template ${\bf{tc}}_{i,p,r}^{\left\{{s,a} \right\}}=\left[ {tc_{i,p,r,k=1}^{\left\{{s,a} \right\}},tc_{i,p,r,k=2}^{\left\{{s,a} \right\}},...,tc_{i,p,r,k=Kc_{i,p,r}^{\left\{{s,a} \right\}}}^{\left\{{s,a} \right\}}} \right]$ describes fragments of~the~reference signatures in the~partition $\left( {p,r} \right)$ of~the~user $i$, associated with the~signal $a$ ($x$ or $y$), created on the~basis of~the~signal $s$ (velocity or pressure), where:

\begin{equation}
\label{eq_n5}
tc_{i,p,r,k}^{\left\{s,a \right\}}=\frac{1}{J}\sum\limits_{j=1}^J {a_{i,j,p,r,k}^{\{s\}}}.
\end{equation}

After determination of~the~templates ${\bf{tc}}_{i,p,r}^{\left\{{s,a} \right\}}$, parameters of~the~fuzzy system for evaluating the~similarity of~the~test signatures to the~reference signatures are determined.

\subsubsection{Determination of~the~parameters of~the~fuzzy system to evaluate the~similarity of~the~test signatures to the~reference signatures}
\label{sec:Wyznaczenie_parametrów_elastycznego_klasyfikatora_rozmytego}

The test signature verification is based on the~answers of~the~fuzzy system for evaluating the~similarity of~the~test signatures to the~reference signatures. Parameters of~the~system must be selected individually for each user from the~database. Moreover, the~algorithm for signature verification should: (a)~work independently of~the~number of~users (its accuracy should not depend on the~number of~users in the~database), (b)~have the~ability to easily add the~signatures of~new users, (c)~not take into account signatures of~other users in the~training and verification phase of~the~signature. This limits the~use of~known methods e.g. from the~field of~non-linear classification and machine learning (evolutionary or gradient). In this paper we propose a new structure of~the~flexible neuro-fuzzy one-class classifier, whose parameters depend on the~reference signature descriptors. They are determined analytically (not in the~process of~supervised learning) and individually for the~user (his/her reference signatures).

The first group of~parameters of~the~proposed system are the~parameters describing differences between the~reference signatures and the~templates in the~partitions. They are used in the~construction of~fuzzy rules described later (see (\ref{eq_Rules})) and determined as follows:

\begin{equation}
\label{eq_dmax}
dmax_{i,p,r}^{\left\{{s,a} \right\}}=\frac{{{\delta _i}}}{{J\cdot Kc_{i,p,r}^{\left\{{s,a} \right\}}}}\sum\limits_{k=1}^{Kc_{i,p,r}^{\left\{{s,a} \right\}}} {\sum\limits_{j=1}^J {\left| {a_{i,j,p,r,k}^{\{s\}} - tc_{i,p,r,k}^{\{s,a\}}} \right|}},
\end{equation}

\noindent $\delta_i$ is a~parameter which ensures matching of~tolerance of~the~system for evaluating similarity in the~test phase (it is assumed that the~test signatures can be created in less comfortable conditions than the~reference signatures, thus ${\delta_i} \ge 1$). Thus, values of~the~parameters describing differences between the~reference signatures and the~templates in the~partitions of~the~user $i$ take into account the~average similarity of~the~reference signature shape to the~templates in the~partitions.

The second group of~parameters of~the~proposed system are weights of~the~partitions. They are used for evaluation of~the~similarity of~the~test signatures to the~templates of~the~reference signatures of~the~user $i$. A~high value of~the~weight means a~small dispersion of~the~shape signal values from the~template ${\bf{tc}}_{i,p,r}^{\left\{{s,a} \right\}}$ for the~reference signatures of~the~user $i$. A~consequence of~the~high value of~the~partition weight is lower tolerance of~the~system for similarity evaluation in the~test phase.

Determination of~the~weights $w_{i,p,r}^{\left\{{s,a} \right\}}$ starts from determination of~a~dispersion of~the~reference signatures signals. The~dispersion is represented by a~standard deviation. Average standard deviation for all samples in the~partition is determined as follows:

\begin{equation}
\label{eq_sigma}
\bar \sigma _{i,p,r}^{\left\{{s,a} \right\}} = \frac{1}{{Kc_{i,p,r}^{\left\{{s,a} \right\}}}}\sum\limits_{k = 1}^{Kc_{i,p,r}^{\left\{{s,a} \right\}}} {\sqrt {\frac{1}{J}\sum\limits_{j = 1}^J {{{\left( {a_{i,j,p,r,k}^{\{s\}} - tc_{i,p,r,k}^{\{s,a\}}} \right)}^2}}}} .
\end{equation}

With the average standard deviation $\bar \sigma _{i,p,r}^{\left\{{s,a} \right\}}$, normalized values of~the~partition weights are determined:

\begin{equation}
\label{eq_w}
w_{i,p,r}^{\left\{{s,a} \right\}}=1 - {{\bar \sigma _{i,p,r}^{\left\{{s,a} \right\}}} \over {\mathop {\max}\limits_{\scriptstyle p=1,2, \ldots,{P^{\left\{s \right\}}} \atop \scriptstyle r=1,2} \left\{{\bar \sigma _{i,p,r}^{\left\{{s,a} \right\}}} \right\}}}.
\end{equation}

Normalization of~the~weights adapts them for use in the~one-class flexible fuzzy system used for evaluation of~the~similarity of~the~test signatures to the~reference signatures. This evaluation is the~basis for recognition of~signature authenticity.

\subsection{Test phase (verification of~signatures)}
\label{sec:Testing_phase}

At the~beginning of~the~test phase (Fig.~\ref{fig:Algorithm}) the~user: (a)~creates one signature, which will be verified and (b)~claims to be a~specific user from the~database. Then parameters of~the~considered user, stored earlier in the~database, are downloaded and the~signature verification is performed. The~list of~the~parameters is as follows: (a)~trajectories of~the~base signature ${{\bf{x}}_{i,jBase}}$, ${{\bf{y}}_{i,jBase}}$, ${{\bf{v}}_{i,jBase}}$ and ${{\bf{z}}_{i,jBase}}$, (b)~vectors of~allocation to the~sections ${\bf{pv}}_i^{\left\{s \right\}}$ and ${\bf{ph}}_i^{\left\{s \right\}}$, (c)~templates of~the~reference signatures ${\bf{tc}}_{i,p,r}^{\left\{{s,a} \right\}}$, (d)~weights of~the~partitions $w_{i,p,r}^{\left\{{s,a} \right\}}$ ($p=1,2, \ldots,{P^{\left\{s \right\}}}$, $r=1,2$) and (e)~the parameters describing differences between the~reference signatures and the~templates in the~partitions $dmax_{i,p,r}^{\left\{{s,a} \right\}}$.

The first step of~the~verification phase is acquisition of~the~test signature. This signature is pre-matched to the~reference base signature, represented by the~trajectories ${{\bf{x}}_{i,jBase}}$, ${{\bf{y}}_{i,jBase}}$ and the~signals ${{\bf{v}}_{i,jBase}}$, ${{\bf{z}}_{i,jBase}}$. This is done analogously as in the~case of~the~reference signatures in the~training phase (Fig.~\ref{fig:PreProcessing}). Normalized test signature is represented by two shape trajectories: ${\bf{xts}}{{\bf{t}}_i}=\left[ {xts{t_{i,k=1}},xts{t_{i,k=2}}, \ldots,xts{t_{i,k={K_i}}}} \right]$ and ${\bf{yts}}{{\bf{t}}_i}=\left[ {yts{t_{i,k=1}},yts{t_{i,k=2}}, \ldots,yts{t_{i,k={K_i}}}} \right]$. Their structure is analogous to the~shape trajectories ${{\bf{x}}_{i,j}}$ and ${{\bf{y}}_{i,j}}$ of~the~reference signatures used in the~training phase, but they do not have the~index $j$ pointing to the~signature.

The second step of~the~verification phase is partitioning of~the~test signature. As a~result of~partitioning of~the~shape trajectories ${\bf{xts}}{{\bf{t}}_i}$ and ${\bf{yts}}{{\bf{t}}_i}$ their fragments denoted as ${\bf{atst}}_{i,p,r}^{\{s\}}=\left[ {a_{i,p,r,k=1}^{\{s\}},a_{i,p,r,k=2}^{\{s\}}, \ldots,a_{i,p,r,k=Kc_{i,p,r}^{\left\{{s,a} \right\}}}^{\{s\}}} \right]$ are obtained. During the~partitioning the~vectors ${\bf{pv}}_i^{\left\{s \right\}}$ and ${\bf{ph}}_i^{\left\{s \right\}}$ are used. They are determined in the~training phase and their signals indicate sections, in which signals of~the~vectors ${\bf{xts}}{{\bf{t}}_i}$ and ${\bf{yts}}{{\bf{t}}_i}$ should be placed. It was carried out in a similar way in the~training phase.

The third step of~the~verification phase (realized after partitioning) is determination of~the~similarity of~fragments of~the~test signature shape trajectories ${\bf{atst}}_{i,p,r}^{\{s\}}$ to the~templates of~the~reference signatures ${\bf{tc}}_{i,p,r}^{\left\{{s,a} \right\}}$ in the~partition $\left( {p,r} \right)$ of~the~user $i$ associated with the~signal $a$ ($x$ or $y$) created on the~basis of~the~signal $s$ (velocity or pressure). It is determined as follows:

\begin{equation}
\label{eq_dtst}
dtst_{i,p,r}^{\left\{{s,a} \right\}}=\frac{1}{{Kc_{i,p,r}^{\left\{{s,a} \right\}}}}\sum\limits_{k=1}^{Kc_{i,p,r}^{\left\{{s,a} \right\}}} {\left| {atst_{i,p,r,k}^{\{s\}} - tc_{i,p,r,k}^{\{s,a\}}} \right|}.
\end{equation}

After determination of~the~similarities $dtst_{i,p,r}^{\left\{{s,a} \right\}}$, total similarity of~the~test signature to the~reference signatures of~the~user $i$ is determined. A decision on the~authenticity of~the~test signature is taken on the~basis of~this similarity. The~structure of~the~system for evaluation of~the~overall similarity is described in Section~\ref{sec:System_oceny_podobieństwa_podpisu_testowego_do_podpisów_wzorcowych}, evaluation of~the~signature reliability is described in Section~\ref{sec:Weryfikacja_podpisu_testowego} and interpretability aspects of~the~system rules are presented in Section~\ref{sec:Aspects_of_interpretability}.

\subsubsection{Evaluation of~the~overall similarity of~the~test signature to the~reference signatures}
\label{sec:System_oceny_podobieństwa_podpisu_testowego_do_podpisów_wzorcowych}

The system evaluating similarity of~the~test signature to the~reference signatures works on the~basis of~the~signals $dtst_{i,p,r}^{\left\{{s,a} \right\}}$ and takes into account the~weights $w_{i,p,r}^{\left\{{s,a} \right\}}$. Its response is the~basis for the~evaluation of~the~signature reliability. The~proposed system works on the~basis of~two fuzzy rules presented as follows:

\begin{equation}
\label{eq_Rules}
\left\{{\begin{array}{*{20}{c}}
{{R^{\left( 1 \right)}}:\left[ {\begin{array}{*{20}{c}}
{{\bf{IF}}\left( {dtst_{i,1,1}^{\left\{{v,x} \right\}}{\bf{is}}A_{i,1,1}^{1\left\{{v,x} \right\}}} \right){\bf{with}}w_{i,1,1}^{\left\{{v,x} \right\}}{\bf{AND}} \ldots}\\
{\ldots {\bf{AND}}\left( {dtst_{i,{P^{\left\{z \right\}}},2}^{\left\{{z,y} \right\}}{\bf{is}}A_{i,{P^{\left\{z \right\}}},2}^{1\left\{{z,y} \right\}}} \right){\bf{with}}w_{i,{P^{\left\{z \right\}}},2}^{\left\{{z,y} \right\}}}\\
{{\bf{THEN}}{y_i}{\bf{is}}{B^1}}
\end{array}} \right]}\\
{{R^{\left( 2 \right)}}:\left[ {\begin{array}{*{20}{c}}
{{\bf{IF}}\left( {dtst_{i,1,1}^{\left\{{v,x} \right\}}{\bf{is}}A_{i,1,1}^{2\left\{{v,x} \right\}}} \right){\bf{with}}w_{i,1,1}^{\left\{{v,x} \right\}}{\bf{AND}} \ldots}\\
{\ldots {\bf{AND}}\left( {dtst_{i,{P^{\left\{z \right\}}},2}^{\left\{{z,y} \right\}}{\bf{is}}A_{i,{P^{\left\{z \right\}}},2}^{2\left\{{z,y} \right\}}} \right){\bf{with}}w_{i,{P^{\left\{z \right\}}},2}^{\left\{{z,y} \right\}}}\\
{{\bf{THEN}}{y_i}{\bf{is}}{B^2}}
\end{array}} \right]}
\end{array}} \right.,
\end{equation}

\noindent where

\begin{figure}[!pth]
\centering
\includegraphics[scale=1.00, clip]{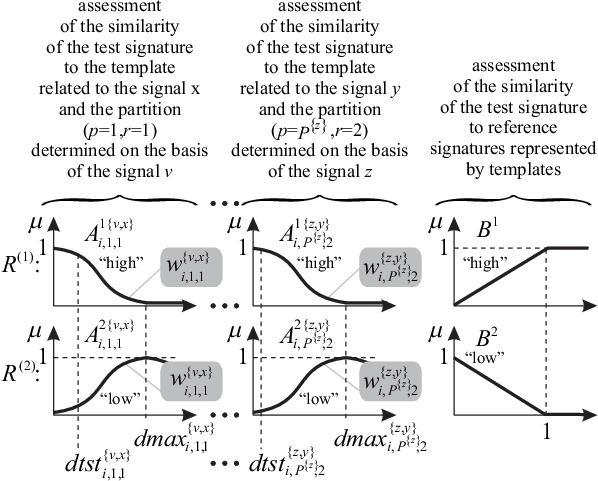}
\caption{Input and output fuzzy sets used in the~rules (\ref{eq_Rules}) of~the~flexible neuro-fuzzy system for evaluation of~similarity of~the~test signature to the~reference signatures.}
\label{fig:Rules}
\end{figure}

\begin{list}{-}{}
\item $dtst_{i,p,r}^{\left\{{s,a} \right\}}$ ($i=1,2, \ldots,I$, $p=1,2, \ldots,{P^{\left\{s \right\}}}$, $r=1,2$, $s \in \left\{{v,z} \right\}$, $a \in \left\{{x,y} \right\}$) are input linguistic variables describing similarity of~the~shape trajectories' fragments ${\bf{atst}}_{i,p,r}^{\{s\}}$ of~the~test signature to the~templates of~the~reference signatures ${\bf{tc}}_{i,p,r}^{\left\{{s,a} \right\}}$. "High" and "low" values taken by these variables are Gaussian fuzzy sets $A_{i,p,r}^{1\left\{{v,x} \right\}}$, $A_{i,p,r}^{2\left\{{v,x} \right\}}$ (see Fig.~\ref{fig:Rules}) described by the~following membership functions:

\begin{equation}
\label{eq_MiG}
{\mu _A}\left( x \right)=\exp \left( {- 1{{\left( {{{x - a} \over \sigma}} \right)}^2}} \right),
\end{equation}

\noindent where $a$ is the~centre of~the~Gaussian function and $\sigma$ is its width (see e.g.~\cite{Rutkowski_2008}). In the used system the value of~the~parameter $a$ for the~rule ${{R^{\left( 1 \right)}}}$ from the rule base (\ref{eq_Rules}) is equal to $0$ and for the~rule ${{R^{\left( 2 \right)}}}$ is equal to the~value of~the~border of~inclusion of~the~reference signatures $dmax_{i,p,r}^{\left\{{s,a} \right\}}$ calculated by the~formula~(\ref{eq_dmax}). The value of~the~parameter $\sigma$ for both rules from~the~rule base (\ref{eq_Rules}) is determined as follows:

\begin{equation}
\label{eq_MiG_sigma}
\sigma ={{dmax_{i,p,r}^{\left\{{s,a} \right\}}} \over {\sqrt {\left| {\log \left( {{\mu _{\min}}} \right)} \right|}}},
\end{equation}

\noindent where ${\mu _{\min}} > 0$ is a~small positive number resulting from the~intersection of~the~Gaussian function (\ref{eq_MiG}) with a~straight line, described by the~equation $\mu \left( x \right)={\mu _{\min}}$, at the~point $\left( {dmax_{i,p,r}^{\left\{{s,a} \right\}},{\mu _{\min}}} \right)$. This approach results from the~specificity of~the~Gaussian function, which tends asymptotically to the~value 0 (this is the~case, in which ${\mu _{\min}}=0$) but never reaches it. Moreover, it is worth noting that the~specific approach to the~fuzzification block of~the~system, and in particular the~use of~the~singleton type fuzzification (see e.g.~\cite{Rutkowski_2008}), allowed us to simplify the~rules notation (\ref{eq_Rules}). The~simplification involves replacing the~names of~linguistic variables by the~names of~the~signals determined using the~formula (\ref{eq_dtst}).

\item $y_i$ ($i=1, \ldots, I$) is output linguistic variable meaning "similarity of~the~test signature to the~reference signatures of~the~user $i$". "High" value of~this variable is the~fuzzy set $B^1$ of~$\gamma$ type (see Fig.~\ref{fig:Rules}), described by the~following membership function (see e.g. \cite{Rutkowski_2008}):

\begin{equation}
\label{eq_MiL}
\mu _{B^1} \left( x \right)=\left\{{\begin{array}{*{20}c}
{\rm{0}} & {{\rm{for}}} & {x \le a} \\
{\frac{{x - a}}{{b - a}}} & {{\rm{for}}} & {a < x \le b} \\
{\rm{1}} & {{\rm{for}}} & {x > b} \\
\end{array}} \right..
\end{equation}

\noindent "Low" value is the~fuzzy set $B^2$ of~$L$ type (see Fig.~\ref{fig:Rules}) described by the~following membership function (see e.g. \cite{Rutkowski_2008}):

\begin{equation}
\label{eq_MiGamma}
\mu _{B^2} \left( x \right)=\left\{{\begin{array}{*{20}c}
{\rm{1}} & {{\rm{for}}} & {x \le a} \\
{\frac{{b - x}}{{b - a}}} & {{\rm{for}}} & {a < x \le b} \\
{\rm{0}} & {{\rm{for}}} & {x > b} \\
\end{array}} \right..
\end{equation}

\noindent In our system value of~the~parameter $a$ for both rules from the~rule base (\ref{eq_Rules}) is equal to $0$ and value of~the~parameter $b$ is equal to $1$.

\item $w_{i,p,r}^{\left\{{s,a} \right\}}$ are weights of~the~partition associated with the~template ${\bf{tc}}_{i,p,r}^{\left\{{s,a} \right\}}$ of~the~user $i$, calculated by the~formula~(\ref{eq_w}). Introduction of~the~weights of~importance distinguishes the~proposed flexible neuro-fuzzy system from typical fuzzy systems.
\end{list}

\subsubsection{Verification of~the~test signature}
\label{sec:Weryfikacja_podpisu_testowego}

In the~proposed method the~test signature is recognized as belonging to the~user $i$ (genuine) if the~assumption ${{\bar y}_i} > ct{h_i}$ is satisfied, where $\bar y_i$ is the~value of~the~output signal of~neuro-fuzzy system described by the~(\ref{eq_Rules}):

\begin{equation}
\label{eq_yi}
{\bar y_i} \approx \frac{{{T^ *}\left\{\begin{array}{c}
{\mu _{A_{i,1,1}^{1\left\{{v,x} \right\}}}}\left( {dtst_{i,1,1}^{\left\{{v,x} \right\}}} \right), \ldots,{\mu _{A_{i,{P^{\left\{z \right\}}},2}^{1\left\{{z,y} \right\}}}}\left( {dtst_{i,{P^{\left\{z \right\}}},2}^{\left\{{z,y} \right\}}} \right);\\
w_{i,1,1}^{\left\{{v,x} \right\}}, \ldots,w_{i,{P^{\left\{z \right\}}},2}^{\left\{{z,y} \right\}}
\end{array} \right\}}}{{\left( \begin{array}{l}
{T^ *}\left\{\begin{array}{c}
{\mu _{A_{i,1,1}^{1\left\{{v,x} \right\}}}}\left( {dtst_{i,1,1}^{\left\{{v,x} \right\}}} \right), \ldots,{\mu _{A_{i,{P^{\left\{z \right\}}},2}^{1\left\{{z,y} \right\}}}}\left( {dtst_{i,{P^{\left\{z \right\}}},2}^{\left\{{z,y} \right\}}} \right);\\
w_{i,1,1}^{\left\{{v,x} \right\}}, \ldots,w_{i,{P^{\left\{z \right\}}},2}^{\left\{{z,y} \right\}}
\end{array} \right\} + \\
+ {T^ *}\left\{\begin{array}{c}
{\mu _{A_{i,1,1}^{2\left\{{v,x} \right\}}}}\left( {dtst_{i,1,1}^{\left\{{v,x} \right\}}} \right), \ldots,{\mu _{A_{i,{P^{\left\{z \right\}}},2}^{2\left\{{z,y} \right\}}}}\left( {dtst_{i,{P^{\left\{z \right\}}},2}^{\left\{{z,y} \right\}}} \right);\\
w_{i,1,1}^{\left\{{v,x} \right\}}, \ldots,w_{i,{P^{\left\{z \right\}}},2}^{\left\{{z,y} \right\}}
\end{array} \right\}
\end{array} \right)}},
\end{equation}

\noindent where

\begin{list}{-}{}
\item $T^*\left\{\cdot\right\}$ is the~weighted t-norm (see \cite{Rutkowski_2008}) in the~form:

\begin{equation}
\label{eq_Tw}
\begin{array}{*{20}{l}}
{{T^*}\left\{\begin{array}{c}
{a_1},{a_2};\\
{w_1},{w_2}
\end{array} \right\}=T\left\{{\begin{array}{*{20}{c}}
{1 - {w_1}\cdot\left( {1 - {a_1}} \right),}\\
{1 - {w_2}\cdot\left( {1 - {a_2}} \right)}
\end{array}} \right\}}\\
{\mathop {\rm{=}}\limits^{{\rm{e}}.{\rm{g}}.} \left( {1 - {w_1}\cdot\left( {1 - {a_1}} \right)} \right)\cdot\left( {1 - {w_2}\cdot\left( {1 - {a_2}} \right)} \right)}
\end{array},
\end{equation}

\noindent where t-norm $T\left\{\cdot \right\}$ is a~generalization of~the~usual two-valued logical conjunction (studied in classical logic), $w_1$ and $w_2 \in\left[0,1\right]$ mean weights of~importance of~the~arguments $a_1, a_2 \in \left[0,1\right]$. Please note that $T^* \left\{{a_1,a_2; 1,1}\right\}=T \left\{{a_1,a_2}\right\}$ and $T^* \left\{{a_1,a_2; 1,0}\right\}=a_1$.

\item $cth_i \in \left[ {0,1} \right]$ - coefficient determined experimentally for each user to eliminate disproportion between FAR and FRR error (see e.g. \cite{Yeung_2009}). The~values of~this coefficient are usually close to $0.5$.
\end{list}

\noindent The formula (\ref{eq_yi}) was established by taking into account the~following simplification resulting from the~spacing of~the~fuzzy sets shown in Fig.~\ref{fig:Rules}:

\begin{equation}
\label{eq_Simplification}
\left\{ {\begin{array}{*{20}{c}}
{{\mu _{{B^1}}}\left( 0 \right) = 0,{\mu _{{B^1}}}\left( 1 \right) = 1}\\
{{\mu _{{B^2}}}\left( 0 \right) = 1,{\mu _{{B^2}}}\left( 1 \right) = 0}
\end{array}} \right..
\end{equation}

\noindent Detailed information about the~system described by the~rules (\ref{eq_Rules}), which allow us to easily derive the~relationship (\ref{eq_yi}) on the~basis of~the~assumption (\ref{eq_Simplification}), can be found e.g. in \cite{Cpalka_IJoGS_2013, Rutkowski_Cpalka_TNN_2005, Cpalka_TNN_2009, Cpalka_NASA_2009, Rutkowski_TNN_2003, Rutkowski_TIE_2012}. It is worth noting that description of~the~considered neuro-fuzzy system given in this paper is sufficient for its implementation. In our previous papers we have dealt in detail with its derivation (it is known in the~literature as flexible neuro-fuzzy system of~the~Mamdani type), gradient and evolutionary learning and applications. The novelty of the algorithm proposed in this paper is a new way of using the system for (a)~assessment of~the~overall similarity of~the~test signature to the~reference signatures and (b)~verification of~the~test signatures. This method does not require a learning step. It allows for a~specific interpretation of~each parameter of~the~system and it takes into account the~specificity of~a~considered problem of~the~dynamic signature verification.

\subsubsection{Aspects of~interpretability}
\label{sec:Aspects_of_interpretability}

In the~literature one can find the~conditions for the~readability of~the~fuzzy systems rules. For example, in \cite{Gacto_2011} 4 levels of~interpretability are shown: Q1: complexity at the~rule base level (it takes into account number of~rules and number of~conditions), Q2: complexity at the~level of~fuzzy partitions (it takes into account number of~membership functions, number of~features or variables), Q3: semantics at the~rule base level (it takes into account rules fired at the~same time), Q4: semantics at the~fuzzy partition level (it takes into account completeness or coverage, normalization, distinguishes ability and complementarity). It should be emphasized that the~rules of~the~form~(\ref{eq_Rules}) meet all the~defined levels.

Moreover, it is worth noting that in the~proposed method: (a)~all parameters of~the~rules are determined analytically and they have a~specific interpretation, (b)~all rules have the~same form for all users, but they differ in the~values of~the~parameters.

Aspects of~interpretability used in the~method are based on our previous experience with interpretability of~rule-based systems (see e.g. \cite{Cpalka_TNN_2009, Cpalka_Lapa_Przybyl_Zalasinski_NC_2014}).

\section{Simulation results}
\label{sec:Simulation_results}

Simulations were performed using authorial test environment written in C\# for two dynamic signature databases:

\begin{list}{-}{}
\item \textbf{MCYT-100 database.} MCYT-100 database contains signatures of~100 users, acquired using a~digitizing tablet. The set of~each user contains 25 genuine signatures from one signature contributor and 25 skilled forgeries from five other contributors.

\item \textbf{BioSecure database.} Simulations were performed using commercial DS2 Signature database which contains signatures of~210 users. The~signatures were acquired in two sessions using a~digitizing tablet. Each session contains 15 genuine signatures and 10 skilled forgeries per person.
\end{list}

In Section~\ref{sec:Przebieg_symulacji} the course of~the~simulation is described in detail. In Section~\ref{sec:Uzyskane_wyniki} tables and figures with the~results of the simulation are published. In Section~\ref{sec:Computational_complexity} computational complexity of~the~proposed algorithm is described. In Section~\ref{sec:Conclusions_from_the_simulations} conclusions from the~simulations are presented.

\subsection{The course of~the~simulation}
\label{sec:Przebieg_symulacji}

For each user from the~MCYT-100 and BioSecure databases we repeated 5 times the~training phase and the~test phase, according to the~block diagram shown in Fig.~\ref{fig:Algorithm}. The~results obtained for all users have been averaged. In each of~the~five performed repetitions we used a~different set of~training signatures. The~described method is commonly used in evaluating the~effectiveness of~the~methods for the~dynamic signature verification, which corresponds to the~standard crossvalidation procedure. The~method of~selection of~the~reference and test signatures for each user was as follows:

\begin{list}{-}{}
\item During the~training phase we used 5 randomly selected genuine signatures of~each signer. These signatures were used to perform the~partitioning, determination of~the~templates and calculation of the parameters of~the~system for evaluating the~test signatures similarity to the~reference signatures.

\item During the~test phase we used 15 genuine signatures and 15 forged signatures for the~MCYT-100 database. For the~BioSecure database we used 10 remaining genuine signatures and all 10 forged signatures.
\end{list}

\subsection{Results of~the~simulations}
\label{sec:Uzyskane_wyniki}

\begin{figure}[!pth]
\centering
\includegraphics[scale=0.9, clip]{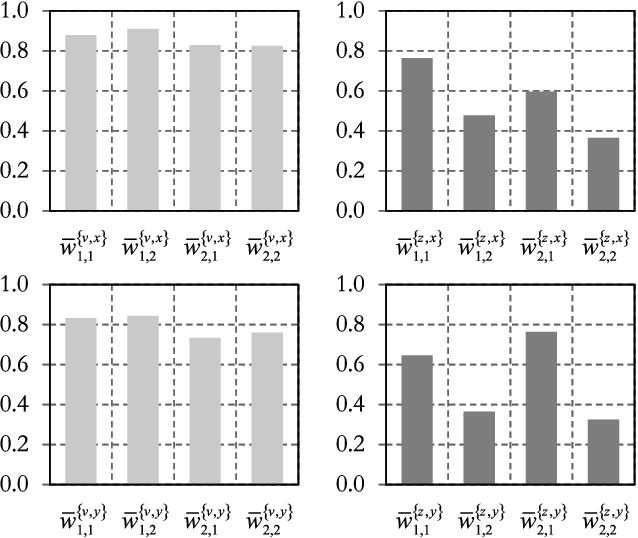}
\caption{Average values of~the~partitions weights created for ${P^{\left\{s \right\}}} = 2$ and ${R^{\left\{s \right\}}} = 2$ (variant of~partitioning for which the~best accuracy has been received), averaged for all users of~the~MCYT-100 should be placed.}
\label{fig:MCYT_Weights}
\end{figure}

\begin{figure}[!pth]
\centering
\includegraphics[scale=0.9, clip]{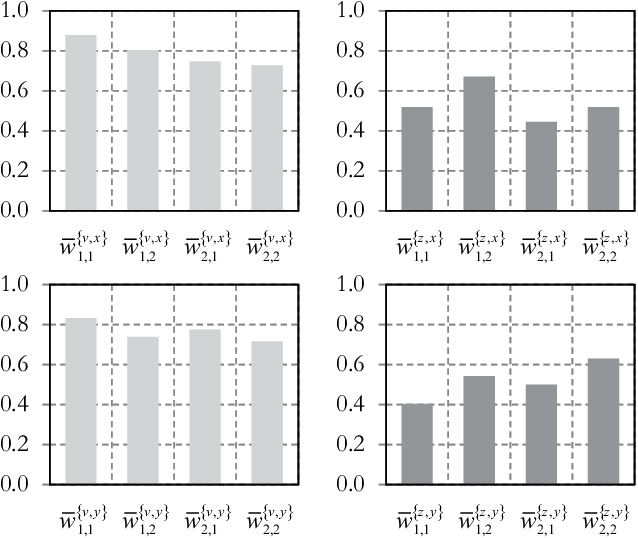}
\caption{Average values of~the~partitions weights created for ${P^{\left\{s \right\}}} = 2$ and ${R^{\left\{s \right\}}} = 2$ (variant of~partitioning for which the~best accuracy has been received), averaged for all users of~the~BioSecure database.}
\label{fig:BS_Weights}
\end{figure}

The results of~the~simulations are presented as follows:

\begin{list}{-}{}
\item Values of~the~errors FAR (False Acceptance Rate) and FRR (False Rejection Rate) for the~database MCYT-100 are presented in Table~\ref{tab_MCYT_FAR_FRR} and for the~BioSecure database are presented in Table~\ref{tab_BS_FAR_FRR}. These errors are used in the~literature to evaluate the~effectiveness of~biometric methods. They have been designated for a~different number of~partitions. Other (less popular) effectiveness measures of~biometric methods may also be used (e.g. the~ones described in \cite{Jain_2008}), but in this case it would be difficult to compare the~obtained results with the~results of~other authors.

\item Comparison of~the~accuracy of~different methods for the~signature verification for the~MCYT-100 database is presented in Table~\ref{tab_MCYT_FAR_FRR_Comparison} and for the~BioSecure database is presented in Table~\ref{tab_BS_FAR_FRR_Comparison}. It is also noteworthy that Table~\ref{tab:Main_features} presents main characteristics of~the~algorithms for the~on-line signature verification based on the regional approach.

\item Weights of~importance of~the~partitions for the~MCYT-100 database are presented in Fig.~\ref{fig:MCYT_Weights} and weights of~importance of~the~partitions for the~BioSecure database are presented in Fig.~\ref{fig:BS_Weights}. Each weight value is the~average value of~the~weights of~all users, determined as follows:

\begin{equation}
\label{eq_w_avg}
\bar w_{p,r}^{\left\{{s,a} \right\}} = \sum\limits_{i = 1}^I {w_{i,p,r}^{\left\{{s,a} \right\}}} .
\end{equation}

\end{list}

\subsection{Computational complexity}
\label{sec:Computational_complexity}

The proposed algorithm does not require high computational complexity (Table~\ref{tab:Computational_complexity}), because it does not require machine learning and does not use any iterative procedures. The~methods proposed in our previous papers (see e.g. \cite{Cpalka_Zalasinski_ESWA_2014,Cpalka_Zalasinski_PR_2014}) also do not require high computational complexity but they contain an~iterative procedure for the purpose of selecting the~so-called border of~inclusion of~the~reference signatures. Elimination of~this procedure from the~proposed algorithm required a~change in the approach to the~evaluation of~the~test signature similarity to the~reference signatures.

Low complexity of~our algorithm is very important in the~verification of~the~test signature genuineness. It enables the system to immediately take into account the signatures of new users.

It has been achieved by eliminating the~machine learning, which usually requires anti-patterns (in the~signature verification they are descriptors of~false signatures, represented by e.g. genuine signatures of~other users). Moreover, in this case a~greater number of~users in the~database causes a~greater variety of~signatures, which provides better learning results. This is a~consequence of~a~more representative learning sequence. In the~proposed algorithm the~number of~users in a~database is not relevant.


\begin{table}
\centering
\caption{Results of~the~simulations performed by our system using the~MCYT-100 database for different number of~sections creating hybrid partitions.}
\label{tab_MCYT_FAR_FRR}
\begin{tabular}{p{1cm}p{1cm}p{2cm}p{2cm}p{2cm}}
\hline\hline
\bf{$P^{\{s\}}$} & \bf{$R^{\{s\}}$} & \bf{Average FAR} & \bf{Average FRR} & \bf{Average error} \\ \hline
\bf{2} & \bf{2} & \bf{5.28} \% & \bf{4.48} \% & \bf{4.88} \% \\ \hline
3 & 2 & 8.69 \% & 5.05 \% & 6.87 \% \\ \hline
4 & 2 & 8.58 \% & 5.73 \% & 7.16 \% \\
\hline\hline
\end{tabular}
\end{table}

\begin{table}
\centering
\caption{Comparison of~the~accuracy of~different methods for the~signature verification for the~MCYT-100 database.}
\label{tab_MCYT_FAR_FRR_Comparison}
\begin{tabular}{p{7.5cm}p{1.5cm}p{1.5cm}p{1.5cm}}
\hline\hline
\bf{Method} & \bf{Average FAR} & \bf{Average FRR} & \bf{Average error} \\ \hline
Methods of~other authors (\cite{Fierrez_2005,Lumini_2009,Nanni_2006, Nanni_Lumini_2006,Nanni_2005,Faundez_2007,Nanni_Lumini_2008, Fierrez_2007,Nanni_2010,Ibrahim_2010}) & - & - & 0.74 \% - 9.80 \% \\ \hline
Algorithm based on Horizontal Partitioning, AHP (\cite{Cpalka_Zalasinski_PR_2014}) & 5.84 \% & 5.20 \% & 5.52 \% \\ \hline
Algorithm based on Vertical Partitioning, AVP (\cite{Cpalka_Zalasinski_ESWA_2014}) & 5.52 \% & 4.87 \% & 5.20 \% \\ \hline
\bf{Our method} & \bf {5.28 \%} & \bf{4.48 \%} & \bf{4.88 \%} \\
\hline\hline
\end{tabular}
\end{table}

\begin{table}
\centering
\caption{Results of~the~simulations performed by our system using the~BioSecure database for different number of~sections creating hybrid partitions.}
\label{tab_BS_FAR_FRR}
\begin{tabular}{p{1cm}p{1cm}p{2cm}p{2cm}p{2cm}}
\hline\hline
\bf{$P^{\{s\}}$} & \bf{$R^{\{s\}}$} & \bf{Average FAR} & \bf{Average FRR} & \bf{Average error} \\ \hline
2 & 2 & \bf{3.36} \% & \bf{3.30} \% & \bf{3.33} \% \\ \hline
3 & 2 & 5.34 \% & 5.56 \% & 5.45 \% \\ \hline
4 & 2 & 9.28 \% & 6.16 \% & 7.72 \% \\
\hline\hline
\end{tabular}
\end{table}

\begin{table}
\centering
\caption{Comparison of~the~accuracy of~different methods for the~signature verification for the~BioSecure database.}
\label{tab_BS_FAR_FRR_Comparison}
\begin{tabular}{p{7.5cm}p{1.5cm}p{1.5cm}p{1.5cm}}
\hline\hline
\bf{Method} & \bf{Average FAR} & \bf{Average FRR} & \bf{Average error} \\ \hline
Methods of~other authors (\cite{Houmani_2009}) & - & - & 3.48 \% - 30.13 \% \\ \hline
Algorithm based on Horizontal Partitioning, AHP (\cite{Cpalka_Zalasinski_PR_2014}) & 2.94 \% & 4.45 \% & 3.70 \% \\ \hline
Algorithm based on Vertical Partitioning, AVP (\cite{Cpalka_Zalasinski_ESWA_2014}) & 3.13 \% & 4.15 \% & 3.64 \% \\ \hline
\bf{Our method} & \bf {3.36 \%} & \bf{3.30 \%} & \bf{3.33 \%} \\
\hline\hline
\end{tabular}
\end{table}

\begin{table}[!t]
\renewcommand{\arraystretch}{1.3}
\caption {The computational complexity of~the~proposed method for the~on-line signature verification using hybrid signature partitioning. In a~bolded step 5, the~computational complexity of~the~proposed method has been reduced in comparison to the~computational complexity of~the~methods proposed by us earlier.}
\label{tab:Computational_complexity}
\centering
\begin{tabular}{|l|l|l|l|}
\hline\hline
\multicolumn{1}{c}{Step number}&\multicolumn{1}{c}{AHP \cite{Cpalka_Zalasinski_PR_2014}}&\multicolumn{1}{c}{AVP \cite{Cpalka_Zalasinski_ESWA_2014}}&\multicolumn{1}{c}{proposed method}\\ \hline
\multicolumn{1}{c}{1} & \multicolumn{1}{c}{$2 R J {K_i}$} & \multicolumn{1}{c}{$2 P J {K_i}$} & \multicolumn{1}{c}{$2 P R J {K_i}$}\\
\multicolumn{1}{c}{2} & \multicolumn{1}{c}{$R J {K_i}$} & \multicolumn{1}{c}{$P J {K_i}$} & \multicolumn{1}{c}{$P R J {K_i}$}\\
\multicolumn{1}{c}{3} & \multicolumn{1}{c}{$R J \left( {{K_i} + 2} \right)$} & \multicolumn{1}{c}{$P J \left( {{K_i} + 2} \right)$} & \multicolumn{1}{c}{$P R J \left( {{K_i} + 2} \right)$}\\
\multicolumn{1}{c}{4} & \multicolumn{1}{c}{$4 R \left( {J + 1} \right)$} & \multicolumn{1}{c}{$4 P \left( {J + 1} \right)$} & \multicolumn{1}{c}{$4 P R \left( {J + 1} \right)$}\\
\multicolumn{1}{c}{\bf{5}} & \multicolumn{1}{c}{$J \left( {R {K_i} + 2} \right) + {n_b}$} & \multicolumn{1}{c}{$J \left( {P {K_i} + 2} \right) + {n_b}$} & \multicolumn{1}{c}{-}\\
\multicolumn{1}{c}{6} & \multicolumn{1}{c}{$R J$} & \multicolumn{1}{c}{$P J$} & \multicolumn{1}{c}{$P R J$}\\
\multicolumn{1}{c}{7} & \multicolumn{1}{c}{$1$} & \multicolumn{1}{c}{$1$} & \multicolumn{1}{c}{$1$}\\ \hline
\multicolumn{4}{p{13.5cm}}{Step number: \textbf{1}.~partitioning of~the~signatures, \textbf{2}.~the~templates generation, \textbf{3}.~determination of~the~distances, \textbf{4}.~determination of~the~weights, \textbf{5}.~determination of~the~border of~inclusion of~the~reference signatures, \textbf{6}.~determination of~fuzzy rules, \textbf{7}.~classification.} \\ \hline
\multicolumn{4}{p{13.5cm}}{$J$ is the~number of~training signatures of~the~user (in our simulations $J=5$), $L_i$ is the~number of~samples of~the~\textit{i}-th signer, $P$ is the~number of~vertical partitions, $R$ is the~number of~horizontal partitions, ${n_b}$ is the~number of~steps of~procedure for determining the~decision boundary for each partition.} \\ \hline\hline
\end{tabular}
\end{table}

\subsection{Conclusions from the~simulations}
\label{sec:Conclusions_from_the_simulations}

The proposed algorithm can be evaluated as follows:

\begin{list}{-}{}
\item For both considered databases it received the~highest accuracy in the~case of~division of~the~signature into two partitions associated with time moments of~signing (initial and final) (see Table~\ref{tab_MCYT_FAR_FRR} and Table~\ref{tab_BS_FAR_FRR}). By analysing the~results presented in Table~\ref{tab_MCYT_FAR_FRR} and Table~\ref{tab_BS_FAR_FRR} it can also be seen that the~accuracy of~the~proposed algorithm does not result from increasing the~number of~sections.

\item It allowed to select the~partitions of~the~signature in which the~reference signatures have been created in the~most stable way. These partitions had the~highest weight value and the~lowest value of~the~parameters describing differences between the~reference signatures and the~templates in the~partitions. It is associated with the~lowest tolerance in evaluation of~similarity of~the~test signatures to the~reference signatures. For the~MCYT-100 database it was the~partition associated with the~shape trajectory $x$, the~initial time moment of~signing and the~high value of~velocity (see Fig.~\ref{fig:MCYT_Weights}). In turn, for the~BioSecure database it was the~partition associated with the~shape trajectory $x$, the~initial time moment of~signing and the~low velocity value (see Fig.~\ref{fig:BS_Weights}). Different results for the~two test databases confirm the~validity of~the~proposed algorithm, which adapts its operation to the~specificity of~the~reference signatures, individually for each user.

\item For the~MCYT-100 database our algorithm promoted partitions associated with the~shape trajectory $x$, because they had higher weight values (see Fig.~\ref{fig:MCYT_Weights}). This shows that for the~MCYT-100 database the~horizontal movement of~the~pen was more characteristic during the creation of~the~reference signatures. In the case of the~BioSecure database the~algorithm promoted partitions associated with the~shape trajectory $x$ for the~signal $v$ and the~partitions associated with both~shape trajectories ($x$ and $y$) for the~signal $z$ (see Fig.~\ref{fig:BS_Weights}). This shows that for the~BioSecure database the~most characteristic for the~users are combinations of~(a)~the horizontal movement of~the~pen and the~velocity signal value and (b)~the vertical movement of~the~pen and both signals value.

\item For both considered databases it promoted partitions associated with the~velocity signal, because they had higher weights values (see Fig.~\ref{fig:MCYT_Weights} and Fig.~\ref{fig:BS_Weights}). It shows that the~most characteristic signal describing dynamics of~the~reference signatures was the~velocity signal.

\item For both considered databases the algorithm worked with good accuracy in comparison to other methods for the~dynamic signature verification (see Table\ref{tab_MCYT_FAR_FRR_Comparison} and Table~\ref{tab_BS_FAR_FRR_Comparison}). Taking into account the~additional benefits of~the~algorithm presented in Table~\ref{tab:Main_features}, it can be said that it is an~interesting method for the~dynamic signature verification.
\end{list}

\section{Conclusions}
\label{sec:Conclusions}

In this paper we proposed the~new algorithm for the~dynamic signature verification using partitioning. The~algorithm uses signals available as a~standard in graphics tablets. It can be also easily adapted to the~specific capabilities of~used hardware, e.g. standard device with a~touch screen, which is not a~graphics tablet. Created partitions are associated with the~areas of~the~signature characterized by: high and low pen velocity and high and low pen pressure on the~graphics tablet surface at initial, middle and final moment of~signing process. The~algorithm assigns to the~partitions weights of~importance, which are used in the~evaluation of~the~similarity of~test signatures to reference signatures. The~evaluation is realized using the~new flexible fuzzy system for the~evaluation of~signature similarity. It uses clear fuzzy rules which allow us to interpret operation of~the~system. Parameters of~the~system are determined analytically. It is realized individually for each user and it does not require signatures of~other users and so-called skilled forgeries. Our algorithm has been tested for two signature databases: MCYT-100 and BioSecure. A~good accuracy of~the~signature verification has been achieved for both databases used; however, it worked significantly better for the~BioSecure database. Achieved accuracy combined with the~additional advantages of~the~proposed algorithm makes it an~interesting solution for identity verification based on the~analysis of~a~handwritten dynamic signature.

The weakness of~the~proposed algorithm is its sensitivity to changes of a~handwritten signature occurring over a~very long period of~time (over a~number of~years). This is a~characteristic of~most biometric methods based on the~behavioural attributes. We are planning to develop a~method which will update the~templates during the~test phase to determine the~trend of~signature changes. Moreover, we are also planning to simplify the~mechanism of~the~similarity evaluation by using only the~most characteristic partition for each user. For this purpose we should, among others, investigate a~dependence between the~accuracy of~the~method and the~number of~used partitions. We should also develop an~effective mechanism for the~selection of~the~partitions.

\section*{Acknowledgment}
The project was financed by the~National Science Centre (Poland) on the~basis of~the~decision no. DEC-2012/05/B/ST7/02138.


\end{document}